\documentclass[]{article}

\usepackage{PRIMEarxiv}
\usepackage{amssymb}
\usepackage{amsmath}
\usepackage{booktabs}
\usepackage{makecell}
\usepackage{lineno}
\usepackage[colorinlistoftodos,prependcaption,textsize=tiny]{todonotes}
\usepackage{color}
\usepackage{soul}
\usepackage{float}
\usepackage{natbib}
\usepackage[T1]{fontenc}
\title{Gaussian Kernel-Based Motion Measurement}

\author{Hongyi Liu, Haifeng Wang \\
    Department of Civil and Environmental Engineering \\
    Washington State University \\
    Pullman \\
    {hy.liu, haifeng.wang}@wsu.edu
}

\begin{document}
\maketitle

\begin{abstract}
The growing demand for structural health monitoring has driven increasing
interest in high-precision motion measurement, as structural information derived
from extracted motions can effectively reflect the current condition of the 
structure. Among various motion measurement techniques, vision-based methods 
stand out due to their low cost, easy installation, and large-scale measurement.
However, when it comes to 
sub-pixel-level motion measurement,
current vision-based methods either lack
sufficient accuracy or
require extensive manual parameter tuning
(e.g., pyramid layers, target pixels, and filter parameters) to reach 
good precision.
To address this issue, we developed a novel Gaussian 
kernel-based motion measurement method, which can extract the motion between 
different frames via tracking the location of Gaussian kernels.
The motion consistency, which fits practical structural conditions,
and a super-resolution constraint,
are introduced to increase accuracy and robustness of our method.
Numerical and experimental validations
show that it can consistently reach high accuracy 
without customized parameter setup for different
test samples.
\end{abstract}

\section{Introduction}
\label{sec1}
Structural motion plays a crucial role in structural health monitoring
as it efficiently reflects the condition of a structure. 
Over the past few decades, structural motion measurement has gained 
widespread attention \cite{yang_review_2021},
triggering the demand for accurate motion measurement.
Based on the sensing technology used,
measurement methods are broadly categorized as contact-based or contactless.
Contact-based methods utilize sensors that physically touch the structure,
such as accelerometers, linear potentiometers, and linear variable
differential transformers (LVDTs).
In contrast,
contactless methods use sensors like cameras and laser scanners that operate
from a distance.
Compared to other methods,
vision-based motion measurement offers notable advantages such as low
cost, rapid deployment, and a large measurement range.

Vision-based motion measurement methods can be categorized into two
major groups: matching-based methods and gradient-based methods
\cite{wang_research_2024, paragios_optical_2006},
Matching-based methods
\cite{rosenfeld_picture_1969,
weng_motion_1989,
xu_review_2018,
wang_research_2024,
huang_survey_2024}
measure motions by
matching a predefined target in two consecutive
video frames and tracking its positional change.
One major approach, known as block matching or template matching,
uses a specific rectangular region of pixels 
(the "template")  as the tracking target
\citep[e.g.,][]{koga_motion_1981,
hashemi_template_2016,
khawase_overview_2017,
singh_improved_2021}.
Block matching methods firstly
define a metric of the brightness distribution to measure
the similarity between two blocks
\citep{hashemi_template_2016,khawase_overview_2017},
and subsequently search for the
most similar block in neighboring frames to determine the motion
\citep{koga_motion_1981,
po_novel_1996,
zhu_new_2000,
zhu_novel_2001,
singh_improved_2021}.
On the other hand,
feature points are another tracking target, which concentrate
on the mathematical description for prominent and robust features of
an image, such as edges, corners, blob patterns, or other learnable features. 
This kind of approach is named as "feature matching"
\citep[e.g.,][]{khuc_computer_2017, xu_review_2018, dong_non-target_2019,
huang_survey_2024}
and the feature descriptor is the core of
feature matching, which developed from handcrafted features
\citep[e.g.,][]{lowe_object_1999,
ke_pca-sift_2004,
leonardis_surf_2006,
hutchison_brief_2010,
tola_daisy_2010,
rublee_orb_2011,
leutenegger_brisk_2011,
hutchison_kaze_2012,
alahi_freak_2012} and learnable descriptors based
on machine learning
\citep[e.g.,][]{yi_lift_2016,
tian_l2-net_2017,
mishchuk_working_2017,
luo_geodesc_2018}.
With these well-developed feature descriptors, feature matching consequently
performs more robustly than block matching to illumination variation
and more complex deformation such as rotation and scaling. However, 
matching-based methods can only obtain pixel-wise coordinates to determine
the position of the target.
Although different interpolation techniques
\citep[e.g.,][]{choi_structural_2011,
sladek_development_2013,
revaud_epicflow_2015}
and correlation
enhancements \citep[e.g.,][]{reddy_fft-based_1996,
zhang_novel_2009,
ye_robust_2020,
hikosaka_image--image_2022} have been developed to
refine the coordinate into sub-pixel level and improve the matching
quality,
the accuracy is still limited to 0.01-0.1 
pixels \citep{sladek_development_2013}, 
ignoring a large portion of motion information embedded in videos
\citep{mas_realistic_2016}.

The second type of vision-based motion measurement methods is gradient-based 
methods, which measure
motions by analyzing the spatial and temporal gradient of video frames.
The earliest approaches include Lucas and Kanade (LK) optical
flow and Horn and Schunck (HS) optical
flow, introduced in 1981 \citep{lucas_iterative_1981,horn_determining_1981}.
These methods assume that the brightness intensity contour will
remain consistent after moving and that the motion is approximately
proportional to the brightness intensity difference compared to the same pixel in the
next frame \citep{chen_modal_2015}.
In other words, motions are explicitly
represented by brightness intensity contours in the image plane. Based on this idea,
physics-based constraints were developed in the
following years to further improve accuracy \citep{sun_secrets_2010,
zimmer_optic_2011,
fortun_optical_2015,
monzon_regularization_2016,
tu_survey_2019}. 
To further improve the accuracy and robustness,
Fleet and Jepson \cite[][]{fleet_computation_1990} proposed phase-based approaches as
another kind of gradient-based method, which is inspired by the principle that
motions in spatial domain correspond to the
phase shift in frequency domain. They
claimed that the evolution of phase contours provides a more robust
approximation
to motion than the brightness intensity contours and introduced phase-based
optical flow. 
Improvements and customizations further enhanced its accuracy and robustness
\citep{gautama_phase-based_2002,chen_modal_2015}, and phase-based optical flow
therefore becomes a powerful tool to extract motions.
While phase-based methods offer strong
potential on sub-pixel accuracy
(e.g., less than 5\% absolute percentage error on amplitude for
20 Hz vibration with amplitude of 0.002-pixel),
inappropriate choice of filter parameters
and target pixels will introduce large motion errors
\citep{diamond_accuracy_2017}. In other words,
the robustness of phase-based optical flow to parameter selection remains a
significant challenge. To this end, researchers have developed
various refinements. For example, Liu et al. \citep{liu_structural_2022}
took advantage of Hilbert transformation for a more stable phase extraction
technique. Miao et al. \citep{miao_phase-based_2022}
exploited the average results of multi-scale filters to reduce the errors.
Miao et al. \citep{miao_phase-based_2023} found the optimal filter
parameters for one kind of specific stripe-like pattern to offer a guidance
to determine the filter. Miao et al. \citep{miao_novel_2023} also developed
a novel marker based on white square patches and provided the
corresponding parameter selection suggestion. 
Zhang et al. \citep{zhang_vibration_2024} replaced the Gabor filter as
log-Gabor filter to further improve the robustness. 
Despite these improvements, 
the optimal parameters for each target brightness intensity pattern must still
be manually selected.

To address this issue,
we developed a novel Gaussian kernel-based motion measurement method
to achieve high accuracy without delicate parameter selection.
In our method, 
frames containing motion information are decomposed
into a set of representative Gaussian kernels,
and the positional variation of kernels is used to determine
the motion within the region of interest.
With the ability to adaptively represent different patterns
our method ensures high accuracy across diverse test samples without the
need for careful parameter selection. 

The remainder of this paper is organized as follows: 
Section 2 introduces our proposed Gaussian kernel-based motion measurement 
method, Section 3 and Section 4 respectively present our numerical and experimental
validations to demonstrate the effectiveness of our method,
and Section 5 summarizes our findings and outlines directions for future
research.

\section{Theory and methods}
\label{sec2}
\subsection{Background}
\label{subsec1}

A video frame, or an image in a broader sense, can be represented by
Gaussian functions \citep{lewitt_alternatives_1992}. 
Such a function
located in a 2D Cartesian coordinate system indicates a specific 
brightness intensity distribution
across the entire plane, which is defined as a "kernel" in this paper.

A Gaussian kernel $G_i(\boldsymbol{X})$ is defined as
\begin{equation} 
    G_i(\boldsymbol{X}) = c_i 
    \text{exp} \left[-\frac{1}{2} 
    (\boldsymbol{X}-\boldsymbol{\mu}_i)^T 
    \boldsymbol{\Sigma}_i^{-1} (\boldsymbol{X}-\boldsymbol{\mu}_i)\right] 
\end{equation}
where $c_i$ is the center brightness of the Gaussian kernel;
$\boldsymbol{X} = (x,y)^T \in \mathbb{R}^2$ determines the 
location of one specific pixel;
$\boldsymbol{\mu}_i = (\mu^i_x,\mu^i_y)^T \in \mathbb{R}^2$ represents the 
kernel center to indicate the kernel position;
and $\boldsymbol{\Sigma}_i = 
\boldsymbol{R}_i\boldsymbol{S}_i\boldsymbol{S}_i^T\boldsymbol{R}_i^T \in 
\mathbb{R}^{2\times2}$ (named covariance matrix)
determines the scale by scaling matrix $\boldsymbol{S}_i\in \mathbb{R}^2$ and 
the orientation by rotation matrix $\boldsymbol{R}_i\in \mathbb{R}^2$
with the angle $\theta_i$, whose expressions are
\begin{equation}
    \boldsymbol{S}_i = \begin{bmatrix} S_x^i&0\\0&S_y^i \end{bmatrix}
\end{equation}
\begin{equation}
    \boldsymbol{R}_i = \begin{bmatrix} 
    \cos\theta_i&-\sin\theta_i\\ \sin\theta_i&\cos\theta_i \end{bmatrix}
\end{equation}

\begin{figure}[H]
    \centering
    \includegraphics[scale=1]{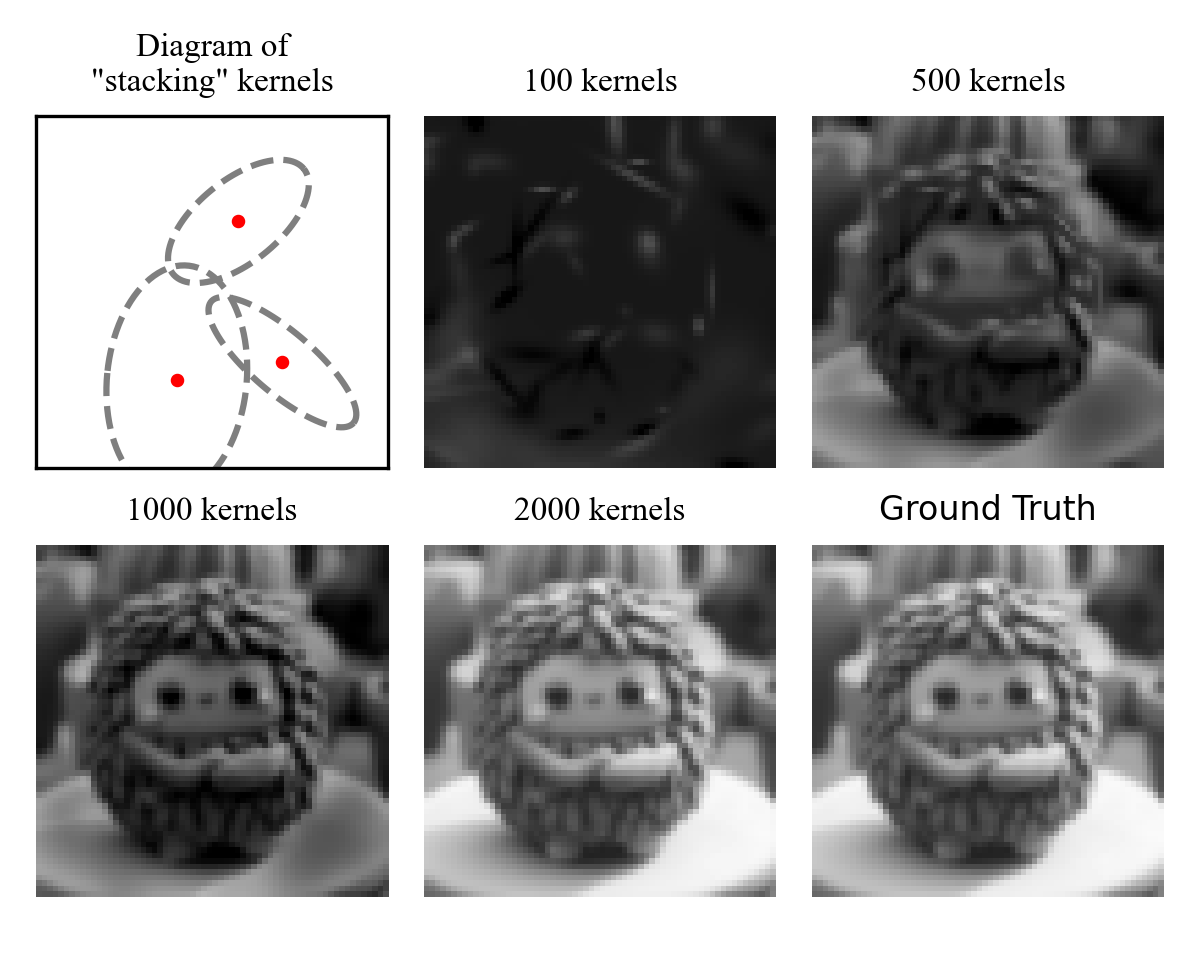} 
    \caption{How to stack various Gaussian kernels to represent an image}
    \label{fig_stack_Gaussians}
\end{figure}

According to the definition, adjusting the Gaussian kernel parameters 
can yield various kernels with a range of intensities, shapes and orientations.
By stacking these Gaussian kernels
as shown in Fig. \ref{fig_stack_Gaussians}, a frame can be 
rendered as the summation 
of Gaussian kernels $\sum^N_{i=1} G_i(\boldsymbol{X})$, and a kernel-based image
representation can be achieved as:
\begin{equation}
    I(\boldsymbol{X}) = \sum^N_{i=1} G_i(\boldsymbol{X}) + E(\boldsymbol{X})
    \label{eq:gaussian_based_representation}
\end{equation}
where $I(\boldsymbol{X})$ is the target frame intensity on a pixel location;
$N$ is the total kernel number;
and $E\left(\boldsymbol{X}\right)$ is the 
representation error.

\subsection{Motion extraction}
\label{subsec2}

To extract motions embedded in frames by the Gaussian kernels, 
the Gaussian kernel set shown in Eq. (\ref{eq:gaussian_based_representation})
should be determined simultaneously
for two consecutive frames $I(\boldsymbol{X})$ and $I'(\boldsymbol{X})$,
where $I(\boldsymbol{X})$ is the current frame and $I'(\boldsymbol{X})$ 
is the next frame.
In this process, 
the mean absolute value of representation errors
$E(\boldsymbol{X})$ and $E'(\boldsymbol{X})$ for both frames are 
minimized using gradient-based kernel parameter optimization.
Specifically, the alternative-splitting-and-pruning
kernel determination process used in
Kerbl et al. \cite{kerbl_3d_2023} and Wurster et al. \cite{wurster_gabor_2024} is
adopted in our method to optimize the kernel parameters
$\{c_i,\boldsymbol{\mu}_i,\boldsymbol{R}_i,\boldsymbol{S}_i\}$.

To achieve high motion measurement accuracy and robustness,
we made two major contributions: 
1) utilizing motion consistency information;
and 2) developing super-resolution constraint.
These two contributions are elaborated below.

\subsubsection{Motion consistency}
Leveraging Gaussian kernels for representation, object deformation
can be linked to the parameter variation of the representative kernels
\citep{wu_4d_2023}.
Under deformation,
the kernel parameter $\boldsymbol{\mu}_i^{'}$, 
$\boldsymbol{R}_i^{'}$ and $\boldsymbol{S}_i^{'}$ in the next frame 
can be represented as:
\begin{equation}
    \boldsymbol{\mu}_i^{'} = \boldsymbol{\mu}_i + \delta\boldsymbol{\mu}_i, 
    \quad 
    \boldsymbol{R}_i^{'} = \boldsymbol{R}_i + \delta\boldsymbol{R}_i,
    \quad
    \boldsymbol{S}_i^{'} = \boldsymbol{S}_i + \delta\boldsymbol{S}_i
\label{eq:deformation}
\end{equation}
where $\delta\boldsymbol{\mu}_i$, $\delta\boldsymbol{R}_i$; and 
$\delta\boldsymbol{S}_i$ are the parameter variations of a Gaussian kernel
on position, orientation and scale compared to the reference frame,
respectively. 
Thus, 
$\Delta = \{\delta\boldsymbol{\mu}_i,\delta\boldsymbol{R}_i,\delta\boldsymbol{S}_i\}$
establishes an equivalence between object
deformation and Gaussian kernel changes.

As for vibration measurement of civil structures,
a region selected from the
target object can be approximately assumed to be a rigid body, 
and thus the motions in this region are uniform.
In other words, in such a localized region,
\begin{equation}
    \delta\boldsymbol{\mu}_i \approx \boldsymbol{C}, 
    \quad 
    \delta\boldsymbol{R}_i \approx 0,
    \quad
    \delta\boldsymbol{S}_i \approx 0,
    \quad \forall i
\end{equation}
where $\boldsymbol{C} = (d_x,d_y)^T$ is a constant for any kernel $G_i$ at time $t$, 
and thus represents the motion of the selected region. 

\subsubsection{Super-resolution constraint}
\begin{figure}[H]  
    \centering
    \includegraphics[scale=1]{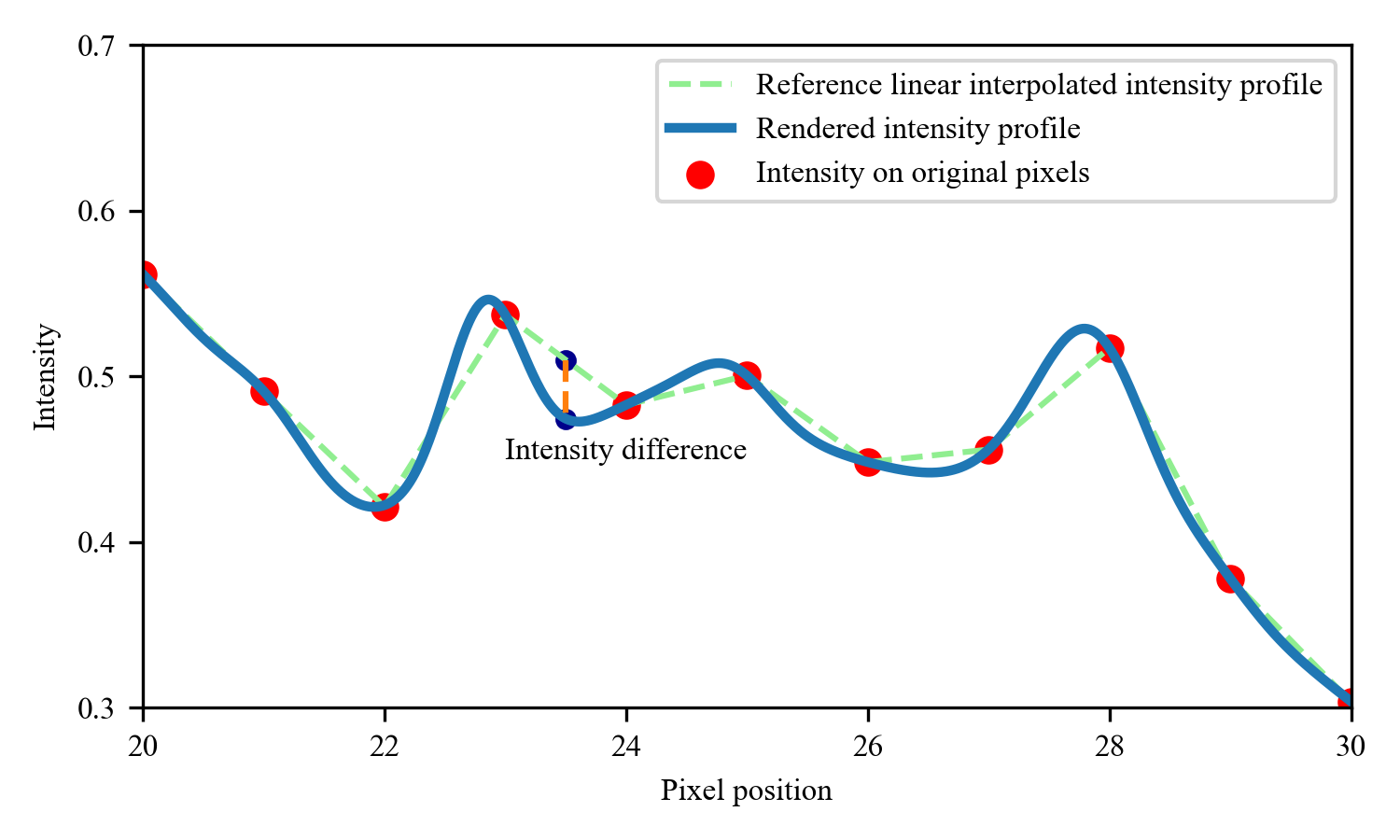}
    \caption{An example of overfitting on one intensity line of a frame}  
    \label{fig_SR_diagram}  
\end{figure}

\begin{figure}[H]  
    \centering
    \includegraphics[scale=1]{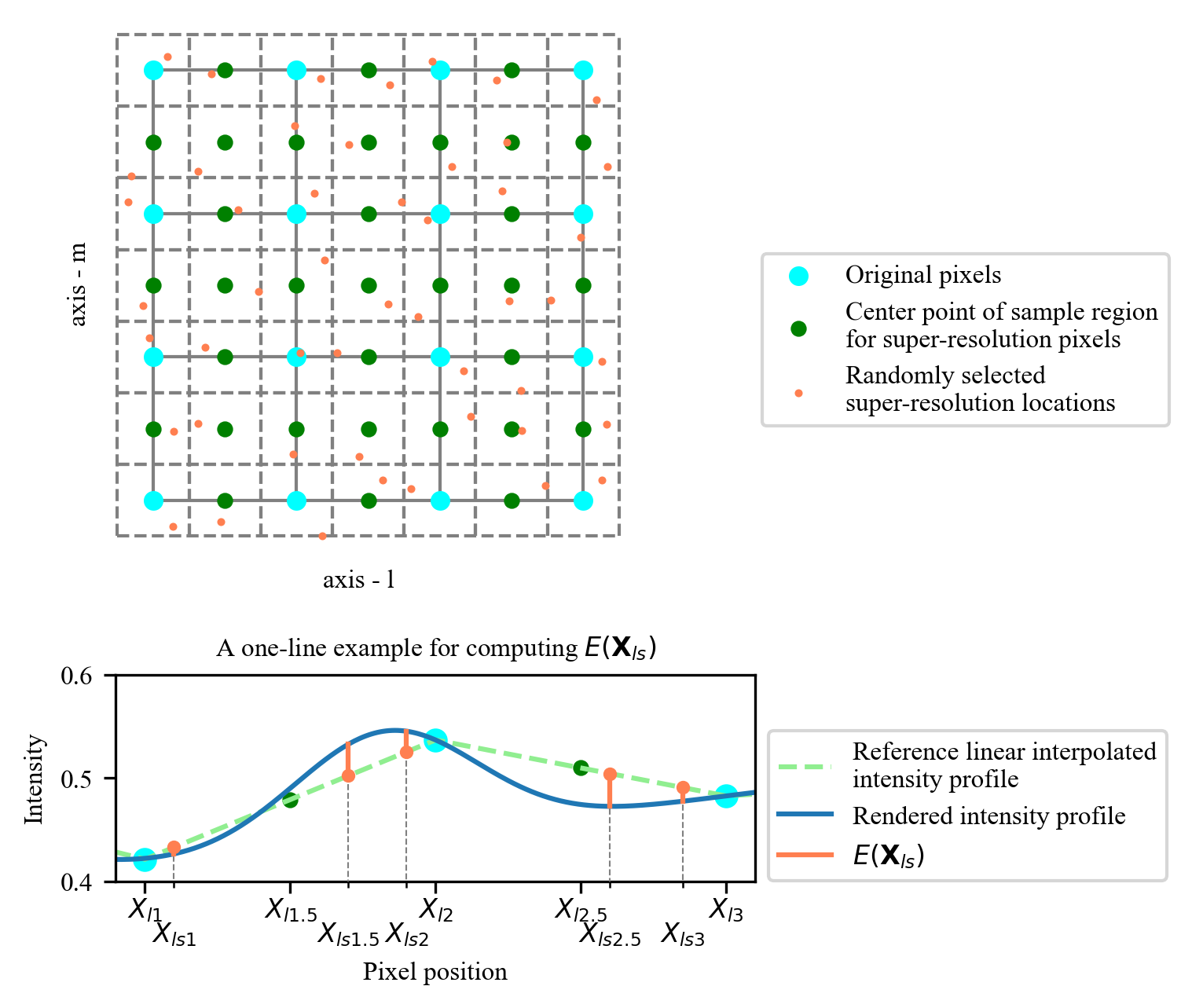} 
    \caption{Sample region for super-resolution locations 
    (Only insert 1 interpolated point between each pair of pixels for display.
    In the following cases, 4 interpolated points were applied)}  
    \label{fig_SR_sample}  
\end{figure}

Because existing kernel parameter optimization targets
\citep[e.g.,][]{kerbl_3d_2023, wurster_gabor_2024}
were primarily designed for
frame generation and visual effects, they concentrate on
intensity differences for each pixel-wise location and thus leads to 5
overfitting issue on inter-pixel regions (Fig. \ref{fig_SR_diagram}).
That is, the brightness between two pixels fluctuates
significantly.
In this case, 
the rendered intensity surface (Eq. \ref{eq:gaussian_based_representation})
will deviate from the underlying image, and
the obtained $\boldsymbol{C}$ will deviate from the actual motion,
leading to incorrectly measured motion. To address this issue,
we developed the super-resolution constraint ($\mathcal{L}_s$) as an auxiliary
term of the loss function in
Eq. (\ref{eq:loss}) to address the overfitting issue.
\begin{equation}
    \mathcal{L} = (1 - w_s) \sum_l \sum_m
   \frac{ 
   \left|E(\boldsymbol{X}_{l,m})\right| +
    \left| E'(\boldsymbol{X}_{l,m}) \right|}{2} 
    + w_s\mathcal{L}_s
\label{eq:loss}
\end{equation}
where $w_s$ is the weight of the super-resolution constraint;
$E$ and $E'$ are respectively the error of current frame and next frame,
$\left| \cdot \right|$ is the absolute error;
$\boldsymbol{X}_{l,m}$ is pixel coordinate;
$l$ and $m$ are respectively the horizontal index and vertical index of 
the pixel;
and $\mathcal{L}_s$ is the super-resolution constraint shown in
Eq. (\ref{eq:ls}) to
regulate the intensity change and reduce overfitting.
\begin{equation}
    \mathcal{L}_s =
    \sum_{(ls,ms) \in\Omega} 
   |E(\boldsymbol{X}_{ls,ms})|
   u(|E(\boldsymbol{X}_{ls,ms})| - \beta)
\label{eq:ls}
\end{equation}
where $\Omega$ is the set of all randomly selected super-resolution locations,
$\boldsymbol{X}_{ls, ms}$ is a super-resolution location in set $\Omega$
corresponding to pixel coordinate $\boldsymbol{X}_{l,m}$;
$u(\cdot)$ is the Heaviside step function;
the error $E(\boldsymbol{X}_{ls,ms})$ is calculated as the difference
between rendered frame and the linearly interpolated
target frame (Fig. \ref{fig_SR_sample});
and $\beta$ is a threshold parameter such that only errors exceeding
$\beta$ are considered.
The super-resolution coordinates $\boldsymbol{X}_{ls,ms}$ are randomly sampled 
for each pixel $\boldsymbol{X}_{l,m}$ at the square region outlined with dashed lines in Fig.
\ref{fig_SR_sample}.
To improve the computational efficiency, only
5\% of the super-resolution coordinates 
are included in the loss function computation for each iteration.

\section{Numerical Validation}
\label{sec3}
Synthetic frames were used to evaluate the performance of Gaussian kernel-based 
motion measurement. Two kinds of synthetic
frames were designed in our validation tests: Gaussian kernel array (GKA) for 
baseline testing and 
general frame with synthetic motions (GFSM) for noise-free testing.

\subsection{Gaussian kernel array (GKA)}
\label{subsec3}

\begin{figure}[H]  
    \centering
    \includegraphics[scale=0.45]{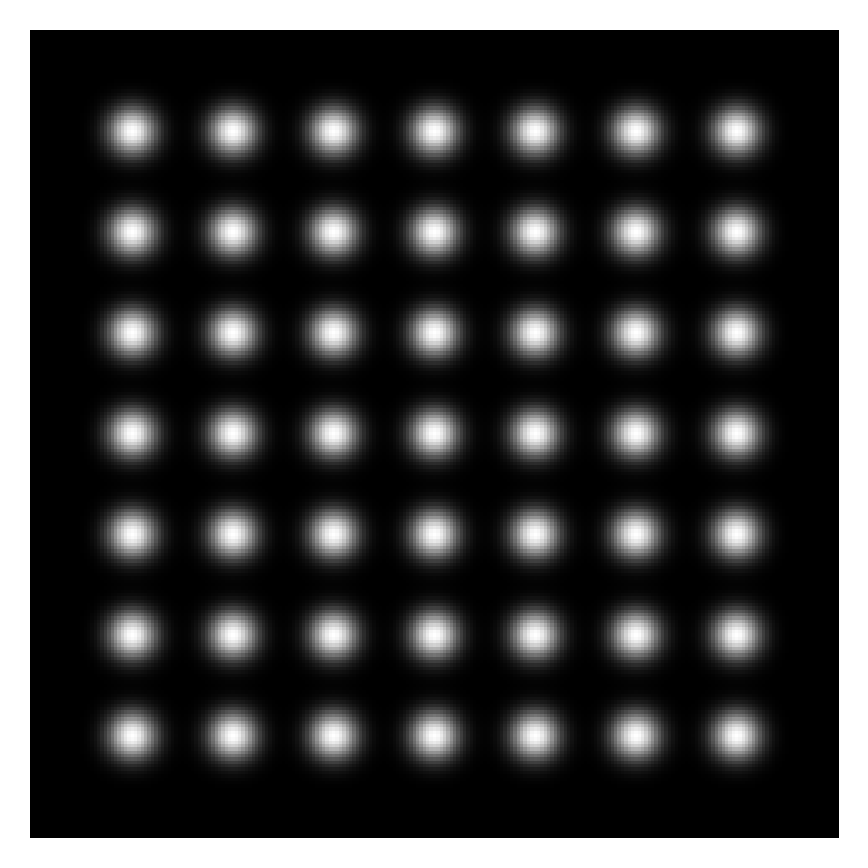} 
    \caption{Gaussian kernel array}  
    \label{fig_Gaussian_array}  
\end{figure}

The Gaussian kernel array (see Fig. \ref{fig_Gaussian_array}) is composed of
multiple Gaussian kernels that are uniformly distributed over a grid-like 
structure. Each kernel shares the same scale and orientation parameters 
($\boldsymbol{S}_i$ and $\boldsymbol{R}_i$), while differing only in position 
($\boldsymbol{\mu}_i$). This design ensures a simple and consistent spatial 
pattern across the array and provides a structured baseline to preliminarily 
explore the accuracy potential of our method. 
Detailed configuration of the frame 
is summarized in Table \ref{tab_Gaussian_array}.

\begin{table}[H]
\centering
\begin{tabular}{ll}
      \toprule
      \textbf{Parameters} & \textbf{Value} \\
      \midrule
      Frame size (resolution) & $241 \times 241$ pixels \\
      Scale $S^i_x, S^i_y$ & 2.52 pixels \\
      Angle $\theta_i$ & $0^\circ$ \\
      Intensity $c_i$ (normalized) & 1.0 \\
      Intensity depth & 8 \& 16 bit \\
      Distance between adjacent kernels & $2.52\times7=17.64$ pixels \\
      \bottomrule
    \end{tabular}
\caption{Parameters for frame of Gaussian kernel array}\label{tab_Gaussian_array}
\end{table}

Although image intensity is inherently continuous,
a camera records the intensity as discrete gray levels determined by its bit depth,
the higher the bit depth, the finer the intensity quantization captured.
To evaluate the accuracy of our method under different intensity quantization
accuracies, 
GKA images with 8-bit and 16-bit were applied.
In addition, sub-pixel translational 
motions were simulated through uniform coordinate shifts of the known
Gaussian kernels
by 0.1, 0.01, and 0.001 pixels in both the horizontal (x-axis) and 
vertical (y-axis) directions.
This allows validation of performance
at multiple motion resolution levels. Combining the two intensity depths
with the three motion levels results in a total of six test cases.
Because of the simple structure of the images, the kernel number
was set as 2000 in GKA tests. Additionally,
for the super-resolution loss term, the loss 
weight $w_s$ was set to 0.33, $\beta$ was set to 0.001 and four
interpolated points were
inserted between each pixel pair for the super-resolution constraint, which is the consistent
configuration for all the following tests.

\begin{figure}[H]  
    \centering
    \includegraphics[scale=1]{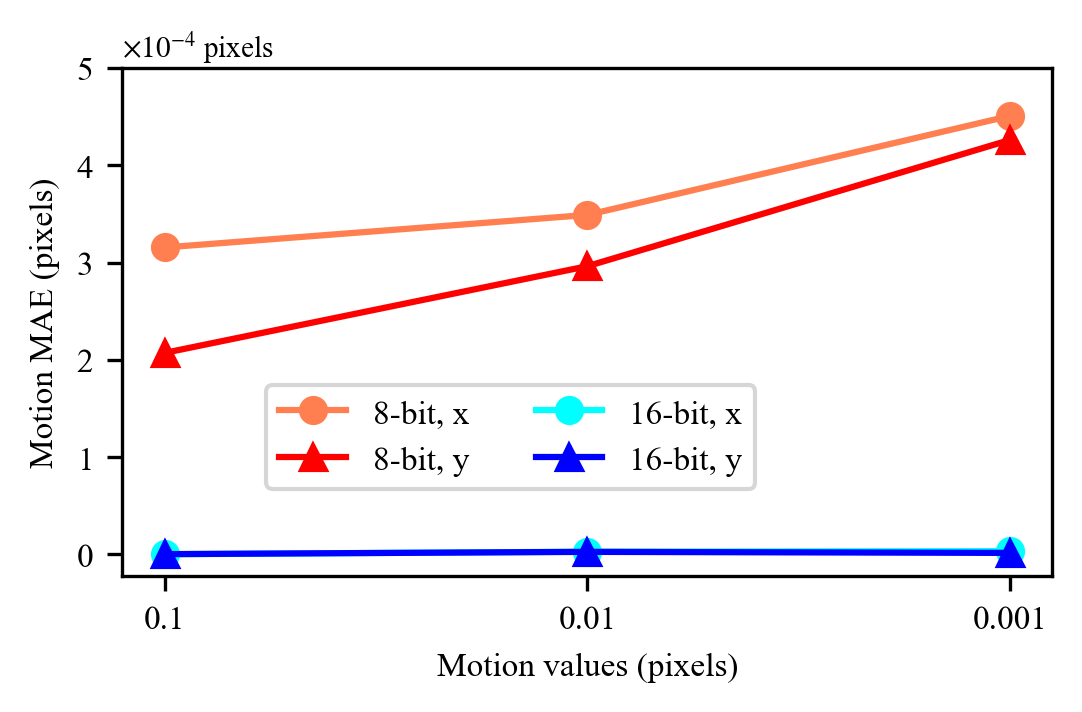}  
    \caption{Motion MAE for GKA}
    \label{fig_res_Gaussian_array}
\end{figure}

To mitigate the influence of initialization, each test case was performed 
using five different random initializations of the Gaussian kernels. The 
resulting mean absolute error (MAE) and standard deviation (STD) are reported 
in Fig. \ref{fig_res_Gaussian_array} and Table \ref{tab_res_Gaussian_array}, 
demonstrating the accuracy of the proposed method under various quantization 
and motion levels.

\begin{table}[H]
\centering
\begin{tabular}{lll}
    	\toprule
    	\textbf{Case} &
    	\makecell{\textbf{MAE(x/y)}\\$\times10^{-4}$pixels} &
    	\makecell{\textbf{STD(x/y)}\\$\times10^{-4}$pixels} \\
    	\midrule
    	GKA-8-0.1 &	1.7/1.5 &	0.2/0.1 \\
    	GKA-8-0.01 &	6.3/6.3 &	0.1/0.1 \\
    	GKA-8-0.001 &	0.6/0.6 &	0.3/0.2 \\
    	GKA-16-0.1 &	0.2/0.2 &	0.0/0.1 \\
    	GKA-16-0.01 &	0.1/0.1 &	0.0/0.0 \\
    	GKA-16-0.001 &	0.1/0.1 &	0.0/0.1 \\
    	\bottomrule
    \end{tabular}
\vspace{0.5em}  
\noindent\parbox{0.6\linewidth}{
\small GKA-($a$)-($b$): $a$ - intensity depth, $b$ - motion value
\\ MAE - mean absolute error
\\ STD - standard deviation.}
\caption{Results of GKA tests}\label{tab_res_Gaussian_array}
\end{table}

For the 16-bit cases, the accuracy reaches the level of $10^{-5}$ pixels, 
corresponding to less than 1\% of all the assigned motions. 
In contrast, the 8-bit cases exhibit satisfactory accuracy with errors 
within 7\% of the respective ground truth 
shifts and only reach $10^{-4}$-pixel level.

\subsection{General frame with synthetic motion (GFSM)}
\label{subsec4}

\begin{figure}[H]  
    \centering
    \includegraphics[scale=1]{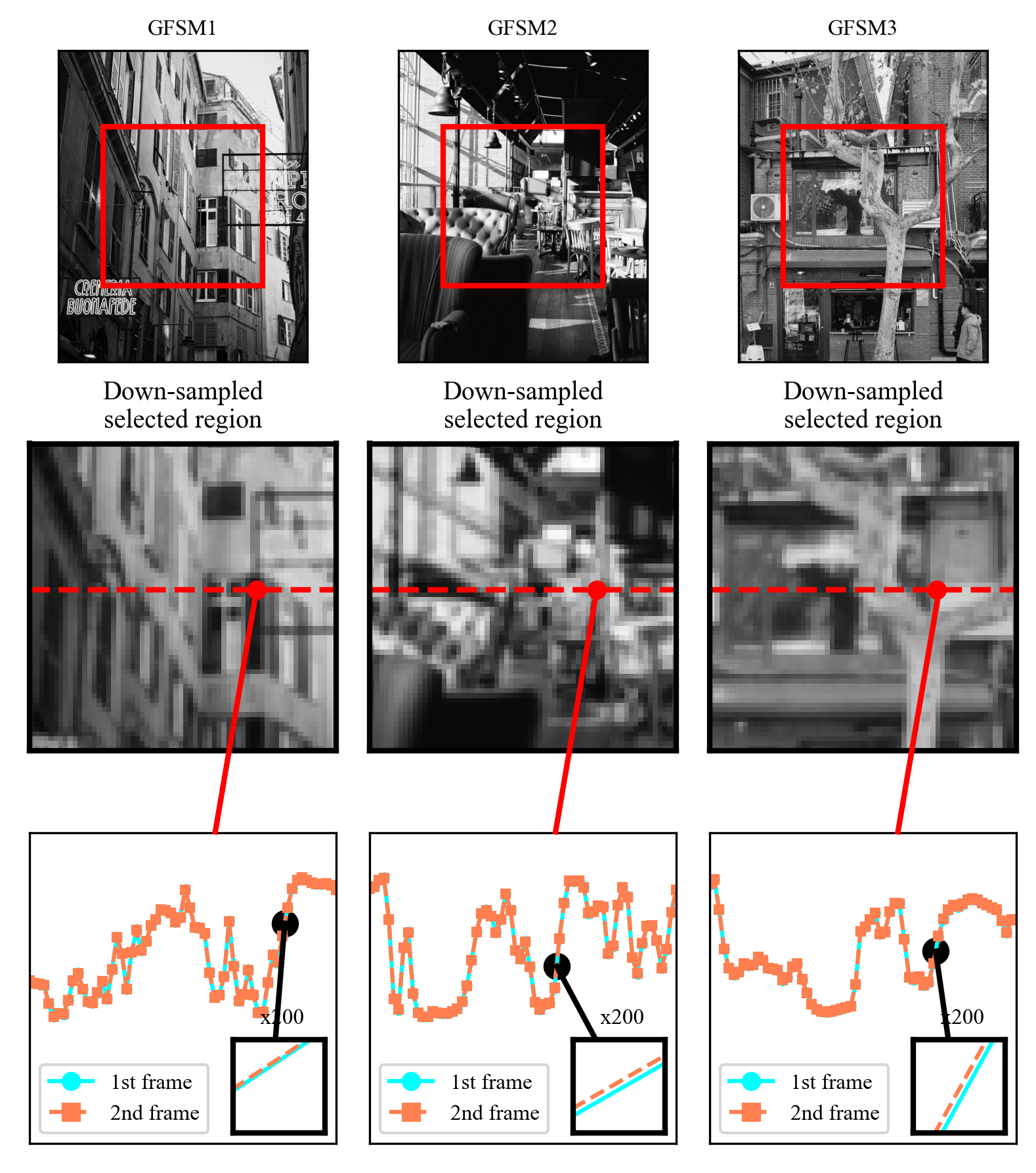} 
    \caption{Test sample images with sub-pixel motions}  
    \label{fig_general_image}  
\end{figure}

To further evaluate the generalization ability of our method, we applied 
it to more complex and diverse image content. As illustrated in 
Fig. \ref{fig_general_image}, three different sample images were 
selected for validation.

Each image was first down-sampled, allowing selected regions to contain
sufficient texture information. From each image, a $64\times64$-pixel 
sub-region was extracted.
These sub-images  serve as test samples under more realistic visual conditions 
compared to GKA setup. Synthetic sub-pixel motions were then 
introduced in these sub-regions under 16-bit intensity resolution, in order to
generate motion-containing frames,
enabling controlled accuracy evaluation under more diverse visual contexts.

To synthesize test frames with controlled sub-pixel motions, we adopted a 
linear interpolation technique described in Miao et al. \cite{miao_novel_2023}.
This method simulates sub-pixel motion by shifting a proportion of the pixel
intensity to the adjacent pixel, following the relationship
$\Delta I = I \cdot d$, where $I$ is the original pixel intensity and 
$d$ denotes the sub-pixel motion along a given direction.

\begin{figure}[H]  
    \centering
    \includegraphics[scale=0.25]{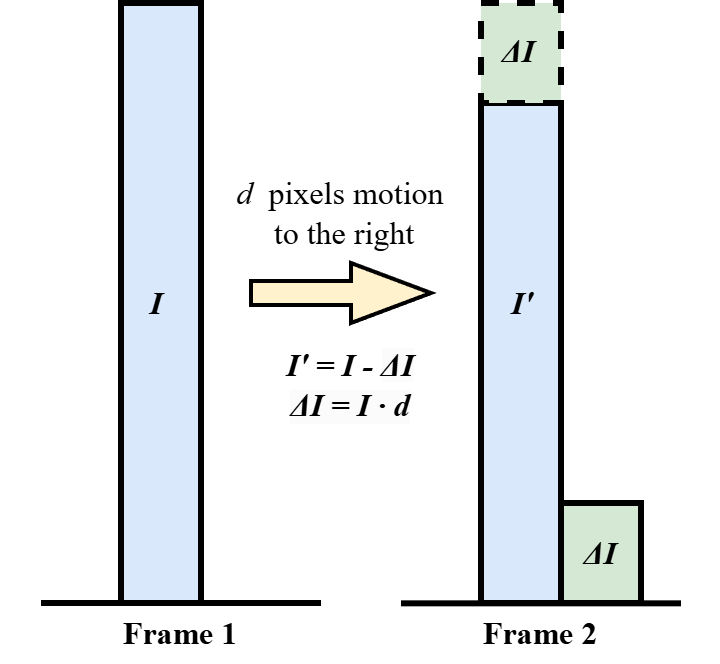} 
    \caption{Diagram of motion synthesis method}  
    \label{fig_intensity_shift}  
\end{figure}

In our implementation, the same motion value was applied simultaneously
in both the horizontal and vertical directions, resulting in diagonal motion
patterns within the synthesized frames. A simplified one-dimensional 
illustration of this interpolation-based motion synthesis is provided in 
Fig. \ref{fig_intensity_shift} to facilitate understanding of the 
pixel-level intensity shift process.

\subsubsection{Kernel number}

The number of kernels used to represent the entire frame has a notable 
impact on motion measurement accuracy. Using too few kernels may lead to 
an under-representation of image content and the lack of fine
details, while an excessive number may cause over-representation and 
result in overfitting.
To explore this trade-off and determine an appropriate kernel number
for each sample image, a dedicated kernel number test was 
conducted. In all test cases, a consistent sub-pixel displacement 
of 0.01 pixels
was applied in both horizontal and vertical directions.

For the super-resolution loss term $\mathcal{L}_s$,  
super-solution pixels were sampled by inserting four interpolated 
points between every 
pair of pixels on the original grid. The associated loss weight $w_s$ 
was set to 0.33, and  
the threshold parameter $\beta$ in $\mathcal{L}_s$ was set to 0.001
to apply a weak linear constraint in the super-resolution loss term.

\begin{figure}[H]  
    \centering
    \includegraphics[scale=1]{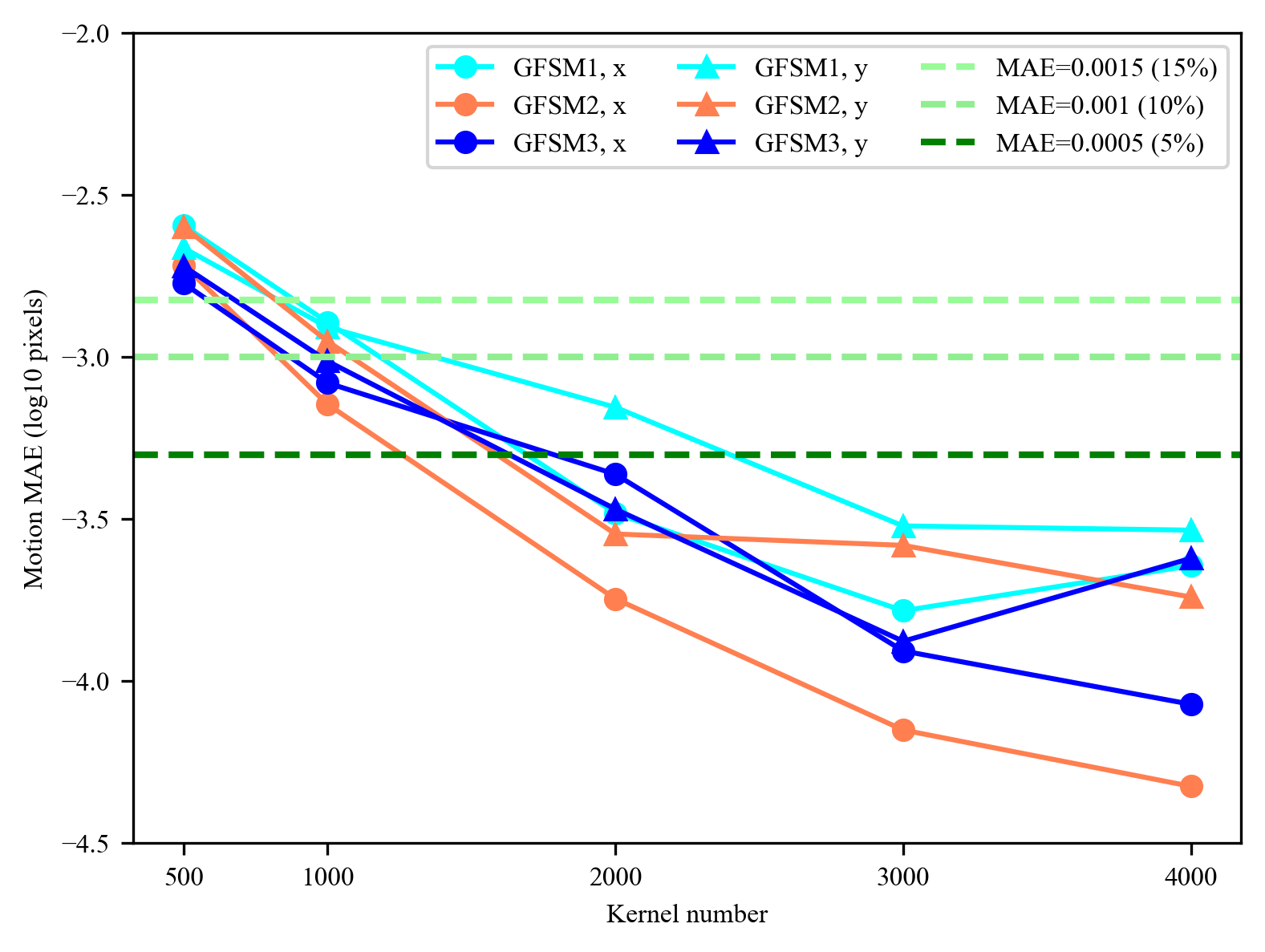} 
    \caption{Motion MAE for GFSM with 0.01-pixel motion 
    on different kernel numbers}  
    \label{fig_res_kernel_num}  
\end{figure}

The resulting motion measurement errors, evaluated in terms of MAE, 
under different kernel 
numbers are presented in Fig.~\ref{fig_res_kernel_num}.
The motion measurement MAEs
generally decrease with increasing kernel number, until reaching an 
optimal point beyond which the error stops decreasing.
This trend reflects
the trade-off between under-representation (insufficient kernels) 
and over-representation (redundant kernels leading to overfitting).

All cases 
achieve a percentage error within 5\% of the ground-truth motion at the
kernel number of 3000. Since the errors at 4000 kernels are not 
consistently better than those at 3000 kernels,
we recommend using 3000 kernels as a general 
setting in our method, without requiring kernel number adjustment across 
different image types. This highlights the robustness and general 
applicability of the proposed configuration.

\subsubsection{Super-resolution constraint loss term}

To evaluate the effectiveness of the super-resolution constraint term 
$\mathcal{L}_s$, a set of controlled tests based on the GFSMs was designed.
Specifically, we compare the motion measurement performance with and 
without incorporating $\mathcal{L}_s$, using GFSMs with various sub-pixel
motion values uniformly applied in both horizontal and vertical directions.

\begin{figure}[H]  
    \centering
    \includegraphics[scale=0.5]{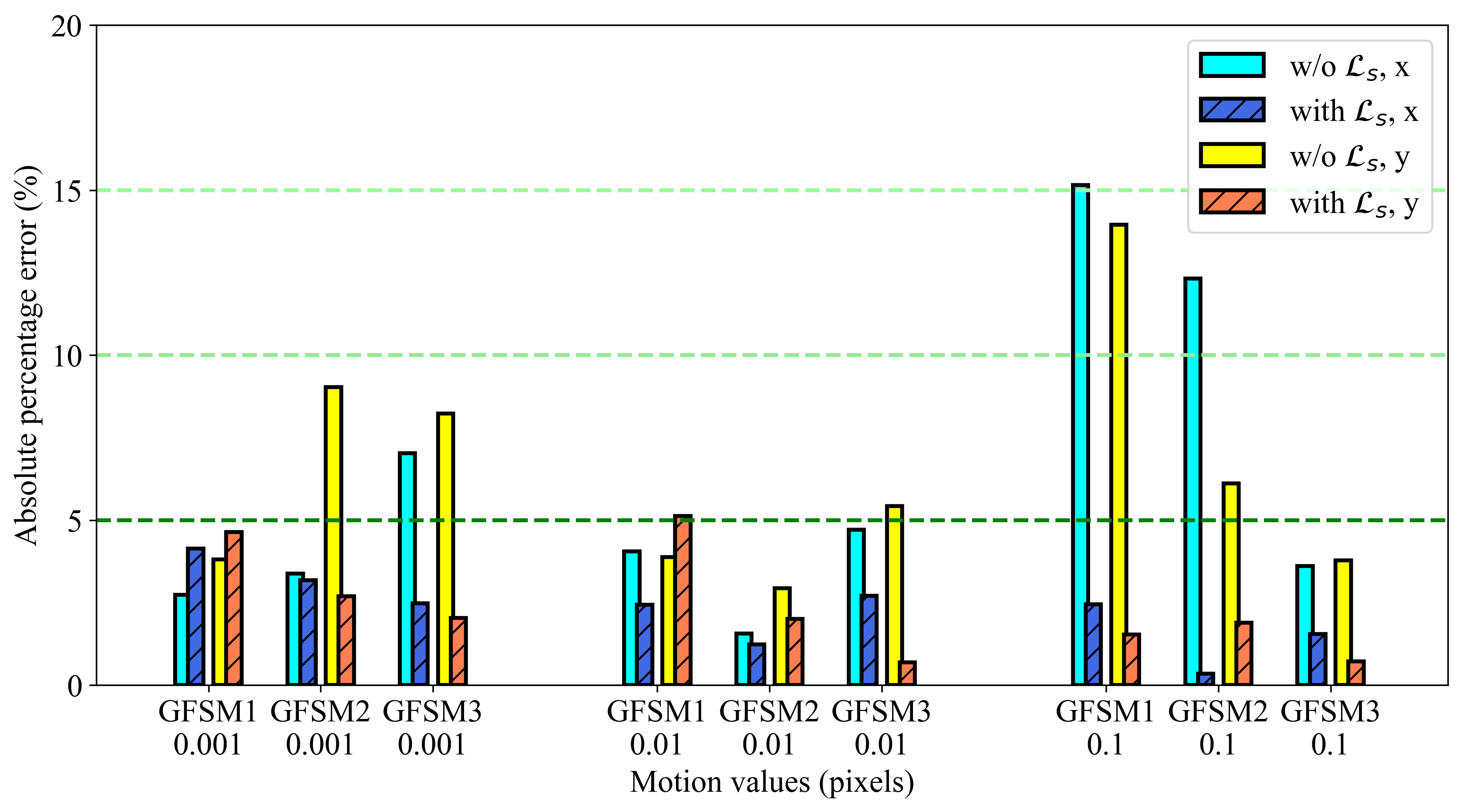} 
    \caption{Motion MAE for GFSM with/without $\mathcal{L}_s$}  
    \label{fig_SR_effect}  
\end{figure}

The kernel number was fixed at 3000 according to the results from 
Section 3.2.1.
The loss-related parameters remained the same as before: 
$w_s = 0.33$, $\beta = 0.001$, and four interpolated points were 
inserted between each pixel pair to construct the super-resolution 
sampling grid. The resulting motion absolute percentage errors for 
each configuration on GFSMs and motions
are shown in Fig.~\ref{fig_SR_effect}.

A significant improvement in motion measurement accuracy is observed
in most test cases when incorporating the super-resolution constraint 
$\mathcal{L}_s$ with an improvement on the percentage error of 
from 5.8\% up to 97.2\%. These results highlight the effectiveness
of the proposed loss term in enhancing sub-pixel motion measurement.

It is also observed that,
in cases with small motion (e.g., GFSM1-0.001 and GFSM1-0.01), 
the effect of the super-resolution constraint is less
significant as compared with large motions. 
This is attributed to the fact that the rendered surface
tends to have smaller over-fitting error for locations nearby
sampling points, i.e., pixel coordinates shown in Fig. \ref{fig_SR_diagram}.
As a result, while the super-resolution constraint reduces the over-fitting
issue, it does not significantly increase the motion estimation accuracy
for small motions.
It should be noted that these less significant exceptions are limited to cases
where the baseline error is already low,
and the percentage error remains within
approximately 5\%, which could be considered as acceptable.
In contrast, for cases
with larger initial errors (e.g., over 10\%), the addition of 
$\mathcal{L}_s$ consistently reduces the error and maintains a stable
performance around the 5\% level, demonstrating its robustness across
diverse motion magnitudes.

\subsubsection{Evaluation across different motion magnitudes}

To further assess the robustness and generalization of the proposed method,
additional GFSM samples were generated with different sub-pixel motion values.
These extended tests were conducted using the same test settings as
in previous evaluations: the kernel number was fixed at 3000, the loss 
weight $w_s$ was set to 0.33, $\beta$ was set to 0.001 and four
interpolated points were
inserted between each pixel pair for the super-resolution constraint.

\begin{figure}[H]  
    \centering
    \includegraphics[scale=0.5]{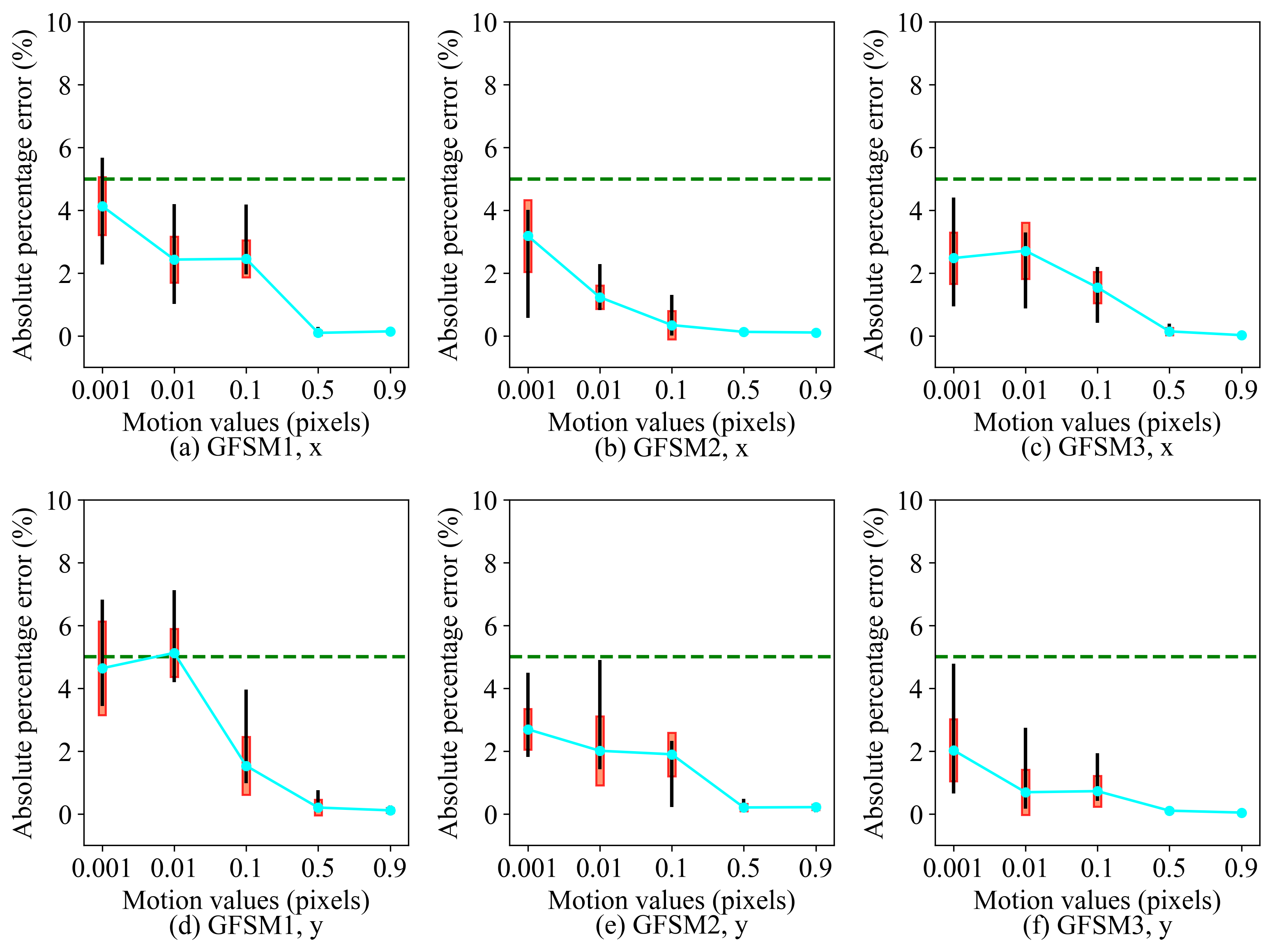} 
    \caption{Motion MAE for GFSM with different motion values}  
    \label{fig_general_res}  
\end{figure}

The results, presented in Fig.~\ref{fig_general_res}, demonstrate that
the proposed method maintains a satisfactory performance across a
broader range of motion magnitudes. The measurement error is lower 
and more consistent at larger motions like 0.5 and 0.9 pixels, but increases 
and becomes less stable as the motions decrease to 0.1, 0.01, and 0.001 
pixels. However, the absolute percentage error 
remains within 5\% for the majority of cases, with a few exceptions staying 
below 10\%, indicating strong robustness to varying motion amplitudes.

\section{Experimental Validation}
\subsection{Experiment setup}
\begin{figure}[H]  
    \centering
    \includegraphics[scale=1]{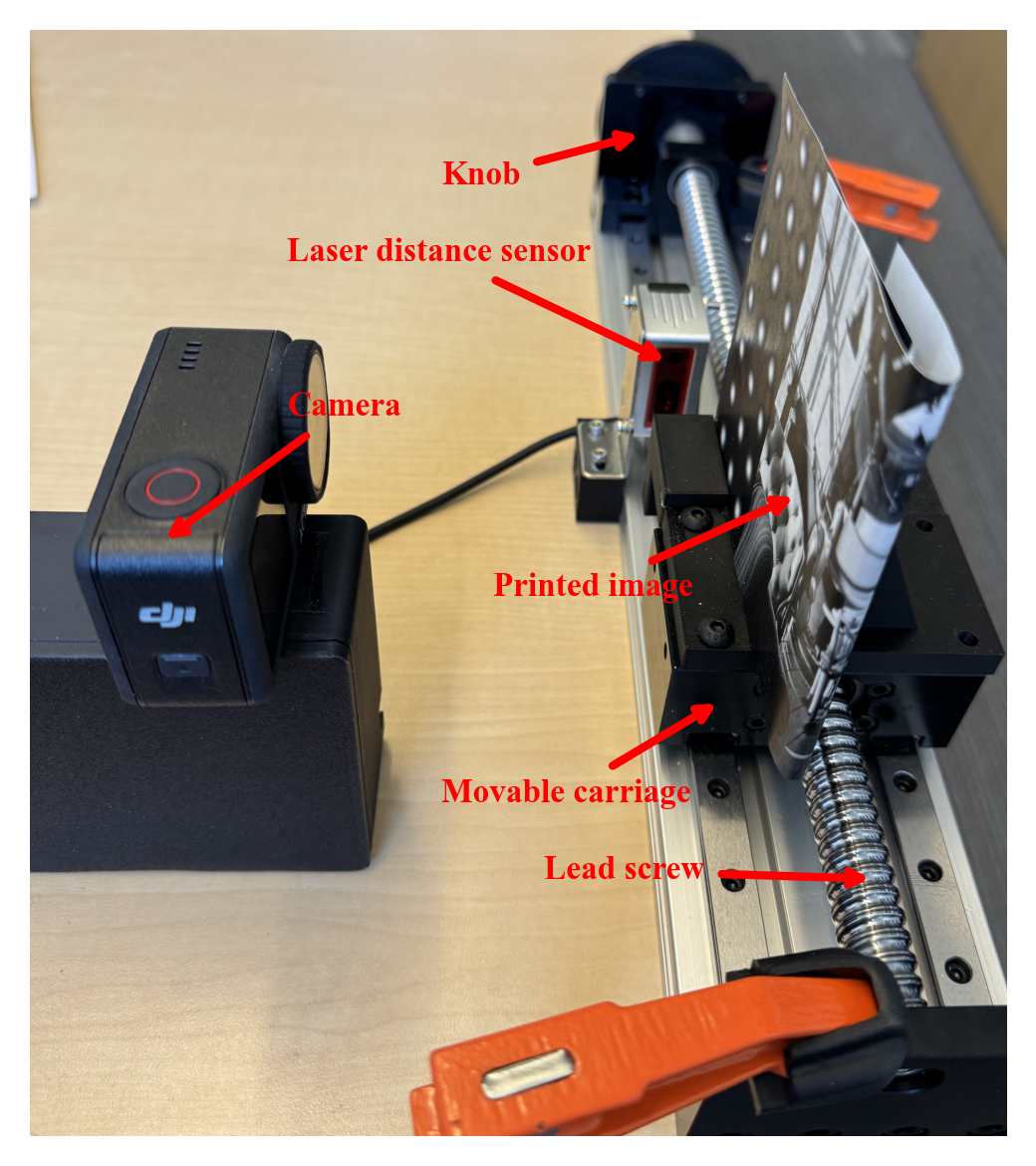} 
    \caption{Experiment setup for the micrometer-driven translation stage}  
    \label{fig_exp_setup}  
\end{figure}

The proposed method was evaluated using a printed target image with
horizontal motion.
The experimental setup is presented in Fig. \ref{fig_exp_setup}.
A lead  screw was placed horizontally to provide accurate linear
translations in the horizontal direction.
A movable carriage
was mounted on the screw and equipped with a clamping mechanism for
holding the target image, which was formed by combining GKA and GFSM2
(as described in Section 3).
Fine translation was 
achieved by turning the knob,
allowing the carriage to move smoothly 
along the screw axis.

\begin{figure}[H]  
    \centering
    \includegraphics[scale=0.5]{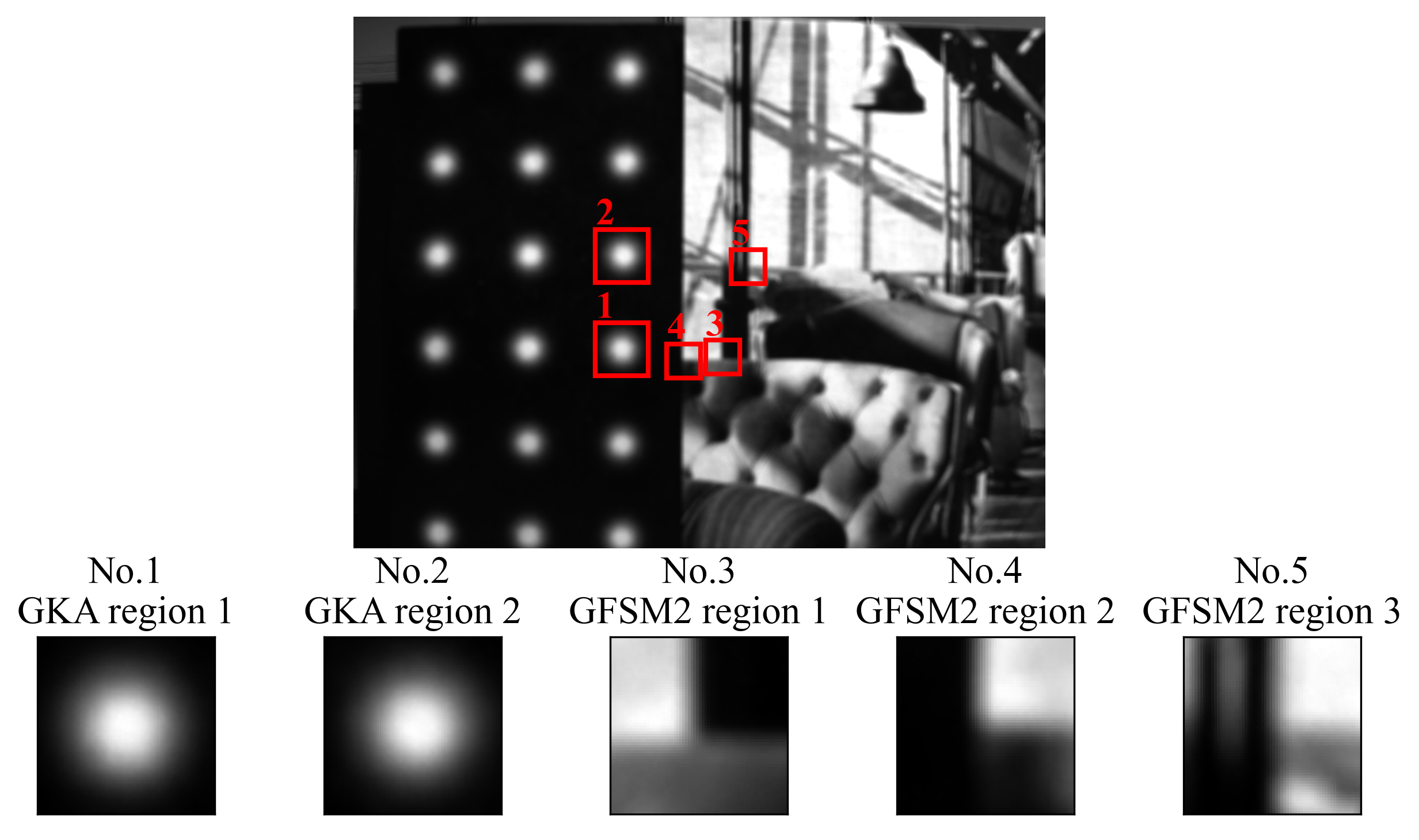} 
    \caption{Selected regions from the frame captured by DJI Osmo Action 5 Pro}  
    \label{fig_select_region}  
\end{figure}

A BOJEK BL-30NZ laser distance sensor was attached
to measure the translation of the carriage as the
ground-truth motion.
In parallel, a DJI Osmo Action 5 Pro camera is 
positioned perpendicularly to the target image plane to capture frames of
the clamped image during the motion. To evaluate the 
accuracy of the proposed method, five regions within the frames were
selected for analysis, including two "GKA regions" and three "GFSM2 regions"
(see Fig.~\ref{fig_select_region}).
Each “GKA region” contains a Gaussian kernel pattern from
GKA with the frame size of $100\times100$ pixels to cover the entire kernel, 
while each "GFSM2 region" includes 
a high-contrast area of GFSM2 and spans $64\times64$ pixels, consistent with
the frame size of the numerical tests in Section 3.
The millimeter-to-pixel ratio is 0.084 and all frames were recorded
as the 14-bit images in the .DNG format.

\subsection{Experimental results}
In total, three ground truth motions were recorded: 0.037, 0.030, and 0.027 mm. 
Given the millimeter-to-pixel ratio, these correspond to motions of
approximately 0.440, 0.357, and 0.321 pixels, respectively indicating
that the motions are in the sub-pixel range.
The optimization configurations followed those used in the GKA and GFSM2 
tests. Given the sufficiently small
millimeter-to-pixel ratio, the intensity variation between two adjacent pixels
can be considered smooth,
allowing the linear relationship between intensity 
variation and sub-pixel motion mentioned in Section 3.2 to remain approximately
valid.

Table~\ref{tab_exp_res} summarizes the MAEs across five random initializations for
each test case. The error level almost remains within 15\% of the actual
motion magnitude, amounting to no more than 0.004 mm and
reaching the minimal error of 0.0008 mm, while the errors on the other direction
(y-axis) are subtle enough compared to the errors on the x-axis.
\begin{table}[H]
\centering
\begin{tabular}{llll}
	\toprule
	\textbf{Ground Truth}\\\textbf{Motion(x/y)} &	0.037/0 & 0.030/0 & 0.027/0\\
	\midrule
	GKA region 1 &	0.0038/0.0001 & 0.0015/0.0000 &0.0029/0.0002\\
	GKA region 2 &	0.0015/0.0000 & 0.0029/0.0002 &0.0032/0.0002\\
	GFSM2 region 1 &	0.0029/0.0002 & 0.0032/0.0002 &0.0008/0.0001\\
	GFSM2 region 2 &	0.0032/0.0002 & 0.0008/0.0001 &0.0031/0.0006\\
	GFSM2 region 3 &	0.0008/0.0001 & 0.0031/0.0006 &0.0036/0.0004\\
	\bottomrule
\end{tabular}
\caption{MAE results for the experiment (unit: mm)}\label{tab_exp_res}
\end{table}

\section{Conclusions}
In this paper, we propose a Gaussian kernel-based motion
measurement method to address the challenging parameter selection
in existing sub-pixel level motion measurement techniques.
By leveraging the representational power of Gaussian
kernels,
our method  achieves stable sub-pixel motion accuracy without
fine-tuned configurations for different test samples.

Through a series of numerical and experimental validations,
we demonstrate that motion can 
be effectively captured by tracking the positional deviation of 
Gaussian kernels across frames. The results validate the 
proposed approach's effectiveness, showing high accuracy 
across diverse image content and motion magnitudes. Key findings
include:

1. The proposed method maintains stable performance across a wide 
range of motion magnitudes (from 0.001 to 0.9 pixels) in the
numerical validations, with percentage errors consistently within 5\%.

2. The inclusion of the super-resolution constraint loss term $\mathcal{L}_s$
significantly improves accuracy by up to 97\% in some cases. Despite a few
exceptions, the overall error remains uniformly within 5\%.  

3. In the experimental validation, our method achieves a MAE
no greater than 0.004 mm, below 15\% of the sub-pixel 
ground-truth motion magnitude. The minimal error can reach 0.0008 mm.

In conclusion, the Gaussian kernel-based motion measurement method presents
a promising advancement in vision-based motion tracking, offering a novel
and effective balance between accuracy and dedicated parameter selection.
The findings of this
study provide a solid foundation for future research and practical applications
in high-accuracy, sub-pixel level structural motion measurement.

\bibliographystyle{unsrt} 
\bibliography{HaifengWangResearchGroup}

\end{document}